\documentclass[letterpaper, 10 pt, conference]{ieeeconf}  

\IEEEoverridecommandlockouts                              

\usepackage{latexsym}
\usepackage{amsmath}
\usepackage{amssymb}
\usepackage{epsfig}
\usepackage{bbm}
\usepackage{caption}
\usepackage{subcaption}

\usepackage{cite}

\usepackage[ruled]{algorithm2e}

\usepackage{verbatim}
\usepackage{hyperref}
\hypersetup{%
  breaklinks,
  bookmarksnumbered=true,
  bookmarksopen=true,
  colorlinks=true,
  linkcolor=black,
  urlcolor=black,
  citecolor=black
}
\usepackage{xcolor}
\usepackage{subfiles}

\begin{document}
\bstctlcite{IEEEexample:BSTcontrol}
%

\title{Learning Risk-Aware Costmaps via Inverse Reinforcement Learning for Off-Road Navigation}




\author{Samuel Triest$^{1}$, Mateo Guaman Castro$^{1}$, Parv Maheshwari$^{2}$, \\ Matthew Sivaprakasam$^{1}$, Wenshan Wang$^{1}$, and Sebastian Scherer$^{1}$ 
\thanks{* This work was supported by ARL awards \#W911NF1820218 and \#W911NF20S0005.}%
\thanks{$^{1}$ Robotics Institute, Carnegie Mellon University, Pittsburgh, PA, USA. \{striest,mguamanc,msivapra,wenshanw,basti\}@andrew.cmu.edu}%
\thanks{$^{2}$ Department of Mathematics, Indian Institute of Technology Kharagpur. parvmaheshwari2002@iitkgp.ac.in}%
}

\maketitle

%
\IEEEpeerreviewmaketitle

\begin{abstract}

The process of designing costmaps for off-road driving tasks is often a challenging and engineering-intensive task. Recent work in costmap design for off-road driving focuses on training deep neural networks to predict costmaps from sensory observations using corpora of expert driving data. However, such approaches are generally subject to overconfident mis-predictions and are rarely evaluated in-the-loop on physical hardware. We present an inverse reinforcement learning-based method of efficiently training deep cost functions that are uncertainty-aware. We do so by leveraging recent advances in highly parallel model-predictive control and robotic risk estimation. In addition to demonstrating improvement at reproducing expert trajectories, we also evaluate the efficacy of these methods in challenging off-road navigation scenarios. We observe that our method significantly outperforms a geometric baseline, resulting in 44\% improvement in expert path reconstruction and 57\% fewer interventions in practice. We also observe that varying the risk tolerance of the vehicle results in qualitatively different navigation behaviors, especially with respect to higher-risk scenarios such as slopes and tall grass. \href{https://drive.google.com/file/d/18LitjcoVRY_s6K-3jK0hRccd7PYAcpsT/view?usp=sharing}{\color{blue} Appendix\footnote[3]{appendix link: tinyurl.com/mtkj63e8}}

\end{abstract}


\section{Introduction}
Navigation in unstructured environments is a major challenge critical to many robotics applications such as off-road navigation, search and rescue, last-mile delivery and robotic exploration. As such, the problem has been widely studied in robotics \cite{team2005stanford, jackel2006darpa, Scherer:2022, arl_stack, borges2022survey, young2021robot}. A key component of navigation in unstructured environments is the capability to take in high-dimensional sensory data and convert it into intermediate representations for downstream planning and control. Typically, this representation takes the form of a costmap, where cells in the costmap describe some notion of the cell's traversability, based on the observed features of that cell. However, determining the mapping between observations and cost is dependent on many factors and is often extremely challenging in practice. 

\begin{figure}
    \centering
    \includegraphics[width=0.87\linewidth]{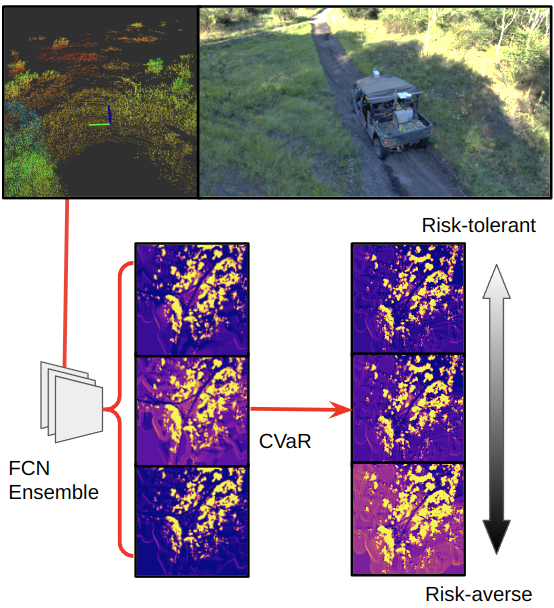}
    \caption{We present an IRL-based costmap learning method that produces risk-aware costmaps from lidar data.}
    \vspace{-0.5cm}
    \label{fig:title_fig}
\end{figure}

Recent work in producing costmaps focuses on directly learning cell costs from map features and expert demonstrations using inverse reinforcement learning (IRL) \cite{wulfmeier2015deep, wulfmeier2017large, wigness2018robot, zhang2018integrating, zhu2020off, lee2022spatiotemporal} and deep neural networks. This is accomplished by training function approximators to predict a mapping from features to cost that best explains the behavior of a large corpora of expert demonstrations. This deep, non-linear representation of the cost function in theory allows for more accurate costmap learning from fewer map features. However, neural networks are often susceptible to domain shift. While this problem is less significant with simple linear combinations of features \cite{abbeel2004apprenticeship, ziebart2008maximum}, increasing the complexity of the function approximator to tens, or hundreds of thousands of parameters can lead to confident mis-predictions for rarely-visited terrain features. In unstructured environments, these mis-predictions can lead to highly undesirable outcomes, such as getting stuck, or damaging the robot. Additionally, standard practice for inverse RL relies on performing value iteration. This approach is both slow, and usually ignores non-holonomic vehicle constraints and dynamics.

In this paper, we describe a costmap learning procedure that extends the standard IRL framework in two key ways:

\begin{enumerate}
    \item Similar to \cite{lee2022spatiotemporal}, we replace the slow value iteration step of traditional IRL methods with MPPI \cite{williams2017model}, allowing for faster training that respects vehicle dynamics.
    
    \item We train an ensemble of cost functions to quantify uncertainty in cost using conditional value-at-risk.
\end{enumerate}

In addition to showing improved results over an occupancy-based baseline, we also validate our costmap learning procedure via navigation experiments on a large all-terrain vehicle and demonstrate that our learned costmaps result in safer navigation than traditional, occupancy-based baselines \cite{arl_stack}. Furthermore, we demonstrate that changing the risk-tolerance of the vehicle results in significant differences in navigation behavior over higher-uncertainty terrain such as slopes and tall grass.

\section{Related Work}

Costmaps are a common representation in off-road navigation tasks due to their ease of use and ability to easily associate cost with spatial locations. As such, a large body of work exists on producing costmaps from sensory data.

Perhaps the most widely-used costmap generation method relies on computing cost from geometric properties of the terrain. The most straightforward form of this approach is to create an occupancy-based representation of the terrain and assign high cost to cells that are occupied with high-height terrain. Kr\"usi et al. \cite{krusi2017driving}, Fankhauser et al. \cite{fankhauser2018robust} and Fan et al. \cite{fan2021step} extend this representation with additional costs derived from geometric terrain features such as curvature and roughness. They all demonstrate that these features allow for improved navigation over rough terrain. 

Recent work in off-road driving has focused on leveraging semantic segmentation approaches to assign traversability costs to different terrain types. Traditional work performs this segmentation in a first-person view (FPV) \cite{maturana2018real, wigness2018robot, jiang2020rellis3d, RUGD2019IROS}. However, it remains challenging to convert this segmentation into a useful representation for navigation, as simple projection is not robust to occlusions or sensor mis-calibration. As such, recent work by Shaban et al. \cite{shaban2022semantic} instead performs semantic segmentation in a bird's-eye view (BEV), using pointclouds from lidar. By first defining a mapping between semantic classes and cost, they are able to train a network to directly output costmaps. While they are able to demonstrate impressive navigation and prediction results, the mapping between semantic classes and cost is both heuristically defined and coarse, limiting the network's ability to generalize between robots or easily adapt online. 

Inverse RL for costmap generation for wheeled vehicles is also a well-explored problem \cite{ratliff2006maximum, ratliff2009learning, bagnell2010learning}. Recent work in generating costmaps for off-road vehicles largely focus on introducing deep neural networks as cost function approximators. Wulfmeier et al. \cite{wulfmeier2017large} trained a deep, fully convolutional network using inverse RL for urban driving scenarios and demonstrate improved performance in generating expert trajectories as compared to a hand-crafted baseline. They also observe an issue in that there are few gradient propagations in rarely-explored areas of the state space, causing their cost networks to perform more poorly on unseen state features. Lee et al. \cite{lee2022spatiotemporal} focus on learning costmaps in a multi-agent highway driving scenario. In addition to replacing the value iteration step of IRL with MPPI \cite{williams2017information}, they also propose regularizing unvisited states to have zero cost to alleviate the costmap artifacts observed by Wulfmeier et al. They additionally add a time dimension to the state-space to account for dynamic agents in their environment. Zhu et al. \cite{zhu2020off} propose a modification to the standard IRL procedure that takes into account an approximation of the kinematics of Ackermann-steered vehicles. They demonstrate that their algorithm results in improved ability to re-create expert behavior on unimproved roads. Zhang et al. \cite{zhang2018integrating} extend the basic IRL framework to include kinematic information. They concatenate information such as velocity and curvature into their intermediate feature representations in order to get costmaps that are able to reflect the current dynamics of their vehicle. Wigness et al. \cite{wigness2018robot} combine semantic segmentation and IRL for a skid-steered robot in off-road scenarios. Unlike previous approaches which focus on extracting costmaps from lidar features, they instead produce costmaps from semantic classes from BEV-projected camera data. Using this feature space, they are then able to learn a costmap using a linear combination of these classes, as well as an obstacle layer. They also demonstrate the capability to quickly update navigation behaviors with a small number of additional demonstrations.

\section{Methodology}
\subsection{Preliminaries}

\subsubsection{MaxEnt IRL}

Maximum Entropy Inverse Reinforcement Learning (MaxEnt IRL) is a popular framework for extracting cost functions for a Markov Decision Process (MDP) from large corpora of human demonstrations \cite{ziebart2008maximum}. MaxEnt IRL builds off of the results shown by Abbeel and Ng \cite{abbeel2004apprenticeship} that one can obtain a policy close to the expert policy by matching feature expectations. However, the baseline IRL problem is ill-posed, as there are an infinite number of cost functions that can explain a set of expert trajectories. Ziebart et al. \cite{ziebart2008maximum} propose using the principle of maximum entropy \cite{jaynes1982rationale} to address the ill-posed nature of unregularized IRL. They show that the gradient for maximizing the likelihood of the expert trajectories $\tau_E$ under the learned reward function with entropy regularization can be computed as Equation \ref{equation:maxent_gradient}.

\begin{equation}
    \label{equation:maxent_gradient}
    \nabla_\theta \sum_{\tau_E \in \mathcal{D}_E} L(\tau_E | \theta) = \tilde{f} - \sum_{s_i} D^L_{s_i} f_{s_i}
\end{equation}

\begin{equation}
    \label{equation:maxent_feat_gradient}
    \nabla_f \sum_{\tau_E \in \mathcal{D}_E} L(\tau_E | \theta) = \sum_{s_i} \left [ D^E_{s_i} - D^L_{s_i} \right ]
\end{equation}

Here, $f_{s_i}$ are the features associated with state $s_i$, $D^L_{s_i}$ is the learner's state visitation distribution, and $\tilde{f}$ is the expert's expected feature counts ($\tilde{f} = \sum_{s_i} D^E_{s_i} f_{s_i}$). Note that the computation of this gradient requires enumeration over the state-space of the MDP.

Performing gradient ascent using this objective has been shown to produce strong results for reward functions linear in state features (of the form $r = \theta^T f$). Wulfmeier et al. \cite{wulfmeier2015deep} extend this formulation by providing a gradient of the MLE objective with respect to state features (Equation \ref{equation:maxent_feat_gradient}). This gradient allows for the training of deep neural networks via IRL (MEDIRL).

\subsubsection{MPPI}

Traditionally, in order to compute state visitation frequencies $D^L_{s_i}$, value iteration is preformed at each gradient step \cite{ziebart2008maximum, wulfmeier2015deep, zhang2018integrating}. For navigation policies in particular, states are generally connected via some pre-defined edge structure, such as a 4 or 8-connected grid \cite{zhang2018integrating}, or an approximation of robot kinematics \cite{zhu2020off}. However, such approaches are relatively slow, and do not obey the true dynamics of the vehicle. We follow an approach similar to \cite{lee2022spatiotemporal} in that we first compute solutions to the MDP in continuous space via a fast MPC algorithm (MPPI), then discretize the resulting solutions ($\tau = x_{1:T}$) into the resolution of the IRL state space.  Given that MPPI can already be thought of as performing importance sampling of the optimal distribution from a proposal distribution, we can also use the existing update weights $\eta$ from MPPI to weight the state visitation frequencies we compute from the MPPI proposal distribution (algorithm in appendix). In order to compute the state distribution of the expert, we can simply set $\eta$ to one. Note that we rely on a position extraction function $p(X) = [x, y]$ to get x-y positions from state. We find that this solution is faster to compute than prior implementations of IRL and respects the kinematic and dynamic constraints of our Ackermann-steered vehicle.






\subsubsection{Conditional Value-at-Risk}

While in theory the implicit regularization of MaxEnt IRL should result in a single weight vector for a given expert dataset, we observe that in practice, especially with neural networks, MaxEnt IRL can converge to meaningfully different solutions. This phenomenon is also noted by Wulfmeier et al. \cite{wulfmeier2017large}. To combat potential uncertainty in feature space, we train an ensemble of function approximators using MEDIRL and combine their predictions using Conditional Value-at-Risk as a risk metric.

While originally used in econometrics applications, there has been interest from the robotics community in using CVaR as a risk metric for reasoning about distributions of cost \cite{majumdar2020should, choudhry2021cvar, fan2021learning, cai2022risk} for its ability to capture more accurately the nuances of long-tailed and multimodal distributions. Intuitively, $\text{CVaR}_\nu$ can be thought of as the the mean value of a distribution, when only considering the portion of the distribution exceeding a given quantile (a.k.a. Value at Risk (VaR)) $\nu \in [0, 1]$ (described formally in Equation \ref{eq:cvar}) \cite{rockafellar2000optimization}.  

\begin{equation}
    \label{eq:cvar}
    \text{CVaR}_\nu = \int^\infty_{f(x) > \text{VaR}_\nu} f(x) p(x) dx
\end{equation}

Note that CVaR can also be used to capture the behavior of the lower tail of the distribution by taking values \textit{below} a given quantile \cite{cai2022risk}. Via a small abuse of notation, we will refer to this mode of CVaR as $\text{CVaR}_{\nu}$ for $\nu \in [-1, 0]$ (Equation \ref{eq:reverse_cvar}). This gives us a range of $\nu \in [-1, 1]$ that we can use to smoothly vary the risk-tolerance of the vehicle. Note that $\text{CVaR}_0$ corresponds to taking the mean of the ensemble (and being neutral to risk), similar to Ratliff et al. \cite{ratliff2009learning}. 

\begin{equation}
    \label{eq:reverse_cvar}
    \text{CVaR}_{-\nu} = \int^{f(x) < \text{VaR}_{1-\nu}}_{-\infty} f(x) p(x) dx
\end{equation}
\subsection{Fast MEDIRL with Uncertainty Estimates}

\begin{figure*}
    \centering
    \vspace{0.2cm}
    \includegraphics[width=0.88\linewidth]{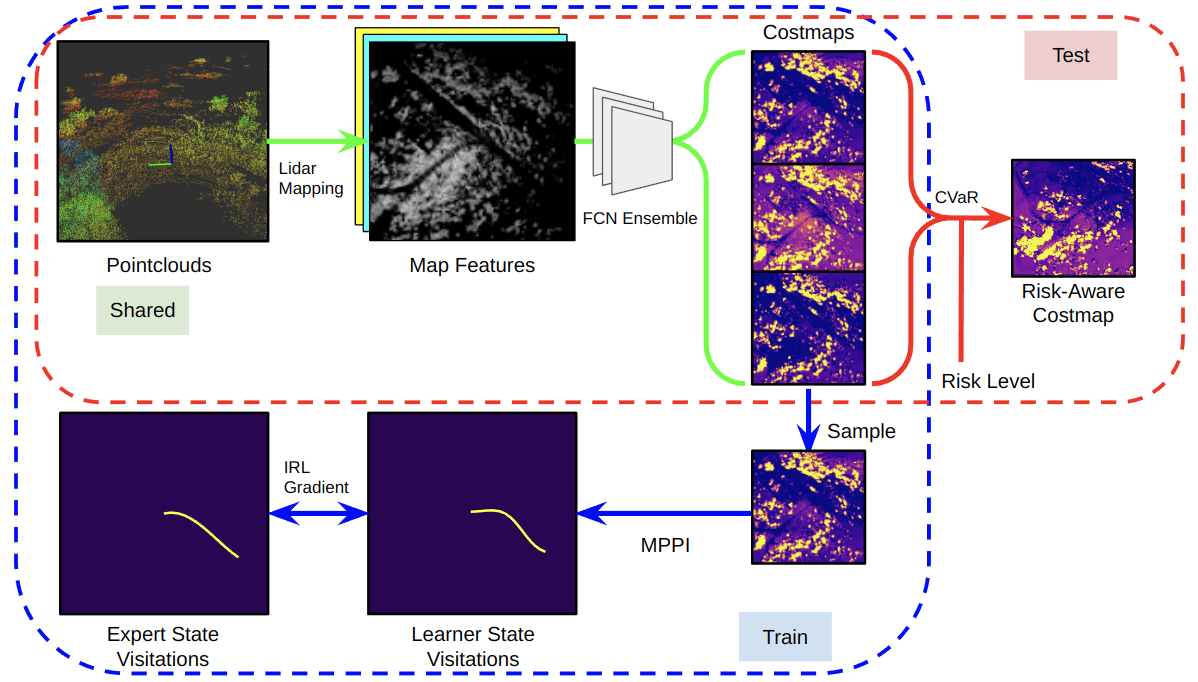}
    \caption{An overview of our algorithm. We first process lidar scans into a tensor of map features. We then use an ensemble of FCNs to create a set of costmaps. In order to train these networks, we can sample from our set of costmaps, solve the resulting MDP with MPPI, and use the resulting state distribution to supervise our networks with inverse RL. At test time, the costmaps can be aggregated together using CVaR to create a costmap associated with a given risk level.}
    \vspace{-0.5cm}
    \label{fig:algo_overview}
\end{figure*}

We present an efficient method for extracting costmaps from expert demonstrations with uncertainty by combining the above methods. An overview of our algorithm is presented in Figure \ref{fig:algo_overview}. At a high level, our algorithm is similar to \cite{wulfmeier2015deep, lee2022spatiotemporal}. However, we train an ensemble of FCNs instead of a single network. This allows us to use ensemble disagreement to estimate uncertainty, similar to prior work \cite{kendall2019geometry, pathak2019self, yu2020mopo, wang2021rough}. In contrast to prior work, we reason over costs. This allows us to to condense our estimates with CVaR, as opposed to using some form of variance. At train time (algorithm in appendix) we construct an MDP by doing the following for each training example $(\tau^E, M_t)$:

\begin{enumerate}
    \item Extract a fixed time-window of expert trajectory $\tau^E_{t:t+H}$ (7.5s or 75 timesteps for our experiments), starting at the time the map features $M_t$ were taken.
    \item Set the first expert state $\tau^E_0$ as the MDP start state.
    \item Set the final expert state $\tau^E_T$ as the MDP goal state.
    \item Obtain a costmap by randomly selecting a costmap predictor from the ensemble and running it on the lidar features at timestep 0.
    \item Set state transition function to be KBM dynamics (equation in appendix).
    \item Set cost function as a weighted combination of the costmap $C_t$ and final-state distance-to-goal (Eq. \ref{eq:cost_function}).
\end{enumerate}

\begin{equation}
    \label{eq:cost_function}
    J(\tau) = \sum_{s_i \in \tau} \Bigl[ C[p(x_i)] \Bigr] + \kappa ||x_T - x_g||_2
\end{equation}

We then solve this MDP via MPPI to obtain the learner's state distribution under the predicted costmap. Using this state distribution, we are then able to update our network weights via the gradient provided in Equation \ref{equation:maxent_gradient}.

An issue noted by Ratliff et al. \cite{ratliff2009learning} is that driving demonstrations may be significantly sub-optimal, especially with respect to homotopy classes around obstacles (e.g. an expert may drive left around a tree, when it is just as good or better to go right). We avoid this issue somewhat by goal-conditioning our MDP based on the final state of a short (7.5s) lookahead window. This also simplifies data collection in that human teleoperators are not required to set and achieve particular goal points.

At test time (algorithm in appendix), we generate a costmap from each FCN and compute a cell-wise CVaR to condense the set of costmaps into a single risk-aware costmap. We can then solve the MDP via MPPI. Unlike train time, however, we can also warm-start the next optimization with the previous solution. In practice, this allows us to run fewer MPPI iterations at test time.

\section{Hardware Implementation}

We tested our algorithms on a Yamaha Viking All-Terrain Vehicle (ATV) which was modified for autonomous driving by Mai et al. \cite{Mai-2020}. The navigation stack was run with a 12th gen Intel i7 cpu and NVIDIA 3080 laptop GPU. A NovAtel PROPAK-V3-RT2i GNSS unit provided global pose estimates as well as IMU data. Additionally, a Velodyne UltraPuck was used to collect pointclouds.

\subsection{State Estimation}

We use Super Odometry \cite{zhao2021super} to provide pose estimates and registered pointclouds, given unregistered lidar scans and IMU data from the NovAtel. 

\subsection{Mapping and Feature Extraction}

Given the registered pointclouds from Super Odometry, we then process the points into a grid map representation \cite{fankhauser2016universal} to perform learning on. This grid map is represented as a dense three-dimensional tensor with two spatial dimensions corresponding to planar position, and the third dimension representing various terrain features of the cell. We represent terrain as an $80m \times 80m$ map at $0.5m$ resolution with 12 features per cell, resulting in a $160 \times 160 \times 12$ tensor. The map is centered at the robot position. We present a simplified description of our lidar mapping algorithm in the appendix, and a description of our lidar features in Table \ref{tab:map_features}.

\begin{table}[]
    \centering
    \begin{tabular}{c|c}
        Feature & Calculation \\
        \hline 
        Height Low & $\min_p[p_z], \quad \forall p | p_z < t_z + k_{overhang}$ \\
        Height Mean & $\frac{1}{|P|} \sum_p [p_z], \quad \forall p | p_z < t_z + k_{overhang}$ \\
        Height High & $\max_p[p_z], \quad \forall p | p_z < t_z + k_{overhang}$ \\
        Height Max & $\max_p[p_z], \quad \forall p$ \\
        Terrain (T) & See appendix \\
        Slope & 0.5 ($|\frac{\partial}{\partial x} T| + |\frac{\partial}{\partial y} T|) $ \\
        Diff & Height high - Terrain \\
        SVD1 & $\frac{\lambda_1 - \lambda_2}{\lambda_1}$ \\
        SVD2 & $\frac{\lambda_2 - \lambda_3}{\lambda_1}$ \\
        SVD3 & $\frac{\lambda_3}{\lambda_1}$ \\
        Roughness & $\frac{\lambda_3}{\lambda_1 + \lambda_2 + \lambda_3}$\\
        Unknown & $\mathbf{1} \left [ |P| = 0 \right ] $ \\
    \end{tabular}
    \caption{List of grid map features, and their calculations given the points in a cell}
    \vspace{-0.5cm}
    \label{tab:map_features}
\end{table}

\subsection{Costmapping}

Given the resulting $[W \times H \times D]$ tensor of map features, we run our ensemble of costmap networks to produce a $[B \times W \times H]$ costmap (where $B$ is the size of the ensemble). Our costmap network is a small, fully convolutional resnet \cite{he2016identity}, which was trained from scratch. We then perform a CVaR calculation for each cell to get a single risk-aware costmap of size $[W \times H]$, given a CVaR value $\nu$.

\subsection{Control}

In order to perform navigation on top of these costmaps, we use a modified version of MPPI \cite{williams2017information}. We observed that sampling noise from an OU process \cite{uhlenbeck1930theory} results in better optimization than the original independent Gaussian noise for our particular application. Our implementation of MPPI minimizes a weighted sum of the costmap and final state distance to a specified goal point (Equation \ref{eq:cost_function}). We optimize trajectories through a kinematic bicycle model (KBM) (equation in appendix) with additional dynamics on throttle and steering. States and derivatives were clamped according to the actuation and safety limits of the vehicle. Similar model and optimization parameters were used for both the inner MPPI optimization of the MEDIRL algorithm and the MPC running on the vehicle (parameters in appendix). To meet the on-board compute and safety constraints of our vehicle, less computationally-heavy optimization parameters, and lower velocity limits were chosen for the on-board MPC.

\subsection{Dataset}

We deploy our algorithm on a corpus of aggressive off-road driving data. Compared to our closest point of comparison, TartanDrive \cite{triest2022tartandrive}, this dataset contains significantly more off-trail scenarios such as pushing through tall grass and bushes, or driving down steep slopes with gravel. Most critically, our dataset contains data from a lidar, allowing for features at a much farther range ($40m$ vs. $10m$ forward in TartanDrive). A more detailed analysis of the driving data is presented in the appendix. In total, we collected roughly one hour of trajectories to train on, and about 15 minutes of separate trajectories to test on. 

\section{Experiments, Results and Analysis}

\subsection{Evaluation of IRL Methods}

Similar to existing literature \cite{zhang2018integrating, wigness2018robot, zhu2020off}, we use modified Hausdorff distance \cite{dubuisson1994modified} as our primary metric to measure the ability of our algorithm to generate costmaps that yield expert behavior (referred to as MHD). For all experiments except the occupancy-based baseline, an ensemble of 16 function approximators were used, and learner trajectories were generated using the mean of the ensemble predictions ($\text{CVaR}_0$). Mean and standard deviation of results are computed over three seeds for each experiment. Results are presented in Table \ref{tab:offline_results}.

\begin{table}[]
    \centering
    \vspace{0.3cm}
    \begin{tabular}{c||c}
        Function Approximator & MHD\\
        \hline
        Occupancy & $3.220 \pm 0.027$ \\
        Linear & $1.847 \pm 0.033$ \\
        Linear Sigmoid & $1.955 \pm 0.046$ \\
        Resnet & $1.973 \pm 0.022$ \\
        Resnet Sigmoid & $\textbf{1.794} \pm \textbf{0.033}$ \\
    \end{tabular}
    \caption{Performance of various function approximators on reproducing expert behavior}
    \vspace{-0.5cm}
    \label{tab:offline_results}
\end{table}

Overall, the IRL-based methods outperformed the occupancy-based baseline by a significant margin. While all IRL-based methods performed similarly, we observe a small, but statistically significant reduction in MHD by using a Resnet. We also observe that ensembles of neural networks exhibit more variation in cost when changing CVaR values (figures in appendix). 

Additionally, we compare the iteration speeds and efficiency of prior IRL methods for offroad driving to our framework. We compare specifically to Zhang et al. \cite{zhang2018integrating} and Zhu et al. \cite{zhu2020off} (Table \ref{tab:implementation_speed}). We are interested in evaluating the time to solve the inner RL loop, as well as the feasibility of these solutions under the motion constraints of our vehicle. Both our method and \cite{zhang2018integrating} were tested on an Intel Xeon CPU. Timing results for \cite{zhang2018integrating} are used from their paper as no source code was available. We compared to their GPU results. Using sampling from actual trajectory rollouts, we are able to achieve faster iteration speed per training sample. Additionally, we are also able to solve the inner RL problem more accurately, as we use the dynamics directly, as opposed to approximating the kinematics.

\begin{table}[]
    \centering
    \vspace{0.3cm}
    \begin{tabular}{c||c|c|c}
        Method & Time per Sample & Kinematics & Dynamics \\
        \hline
        \cite{zhang2018integrating} & 2.0s & No & No \\
        \cite{zhu2020off} & 0.9s & \textbf{Yes} & No \\
        Ours & \textbf{0.3s} & \textbf{Yes} & \textbf{Yes} \\
    \end{tabular}
    \caption{Comparison of properties of different IRL methods. In addition to having the fastest iteration time per sample, our method also respects both the kinematics and dynamics of the vehicle.}
    \label{tab:implementation_speed}
\end{table}

\subsection{Large-Scale Navigation Experiments}

\begin{figure*}
     \centering
     \vspace{0.1cm}
     \begin{subfigure}[t]{0.85\linewidth}
         \centering
         \includegraphics[width=\linewidth]{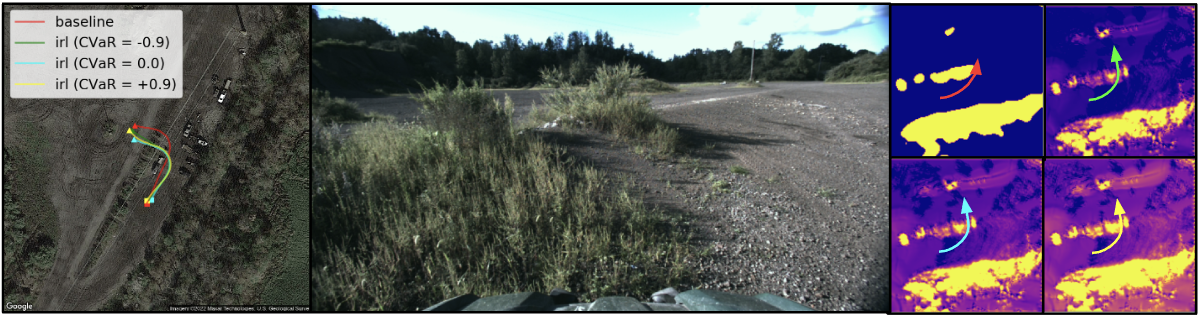}
         \caption{Costmaps learned via IRL allow the robot to cut a corner over some low grass (about 15cm), while the baseline takes a longer route (note that obstacle inflation makes the grass patch appear as an obstacle to the baseline).}
     \end{subfigure}
     \begin{subfigure}[t]{0.85\linewidth}
         \centering
         \includegraphics[width=\linewidth]{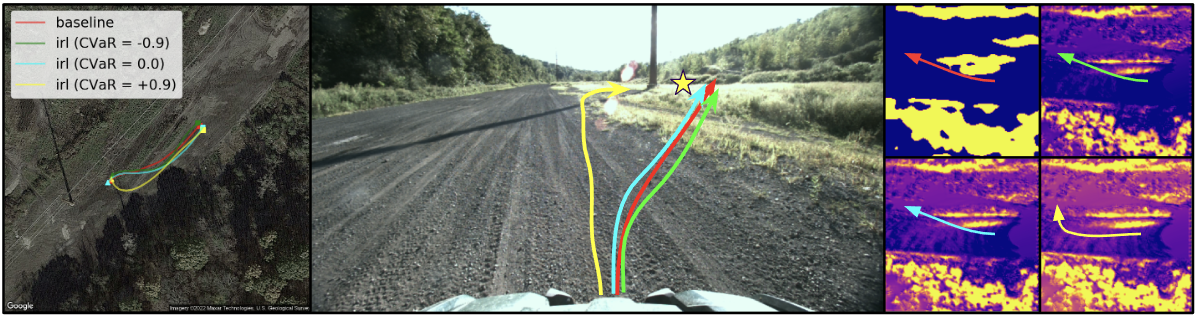}
         \caption{Both the baseline and some IRL costmaps result in navigation over roughly 25cm grass, while IRL with a low risk tolerance (CVaR=0.9), takes a longer route around.}
     \end{subfigure}
     \begin{subfigure}[t]{0.85\linewidth}
         \centering
         \includegraphics[width=\linewidth]{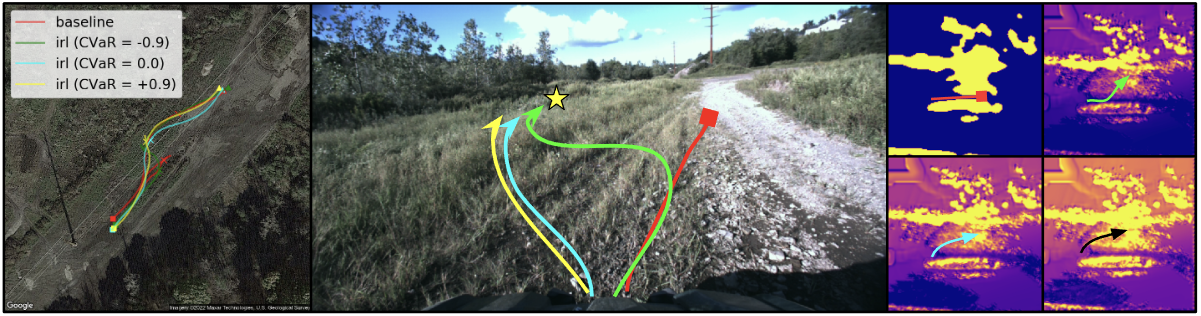}
         \caption{The baseline method is unable to navigate to the waypoint in the tall grass, resulting in an intervention. IRL with a high risk tolerance (CVaR=-0.9) stays on the trail longer, then cuts through a taller grass patch. (black arrow used for clarity in costmap plot for CVaR=0.9)}
     \end{subfigure}
     \caption{Several scenarios that during the navigation run that resulted in different behaviors. Left column: An enlarged BEV of the particular scenario, with trajectories from the individual runs superimposed (start=square, end=triangle, intervention=X, waypoint=yellow star). Middle column: FPV of the given scenario, with the paths from each trial annotated on. Right column: Visualization of the respective costmaps and trajectories for each. (top left: baseline, top right: IRL (CVaR -0.9), bottom left: IRL (CVaR 0.0), bottom right: IRL (CVaR 0.9). }
    \vspace{-0.5cm}
     \label{fig:qualitative_nav}
\end{figure*}

Large-scale navigation experiments were conducted at a testing site near Pittsburgh, PA. The experiments consisted of navigation through a challenging 1.6km course that was defined by a series of GPS waypoints. Given the sensing horizon ($40m$ in all directions) and focus on local planning, waypoints were spaced $50m$ apart. In all experiments, all components of our navigation stack were identical, except for the costmap generation method. As we are interested in evaluating the effect of the costmap on overall navigation performance, we report the number of interventions from a human safety driver as our primary metric. We also report the total distance traveled and average speed, excluding distance traveled while intervening. The safety driver was instructed to intervene in the following two cases:

\begin{enumerate}
    \item If the ATV was going to drive into an obstacle
    \item If the ATV drove past a waypoint without going within a $4m$ radius of it
\end{enumerate}

An intervention consisted of the safety driver tele-operating the vehicle past the point that resulted in the intervention such that a troublesome terrain feature would not result in multiple interventions. Results are provided in Table \ref{tab:intervention_results}. Additional figures are provided in the appendix. Overall, it can be observed all IRL-based methods outperformed the occupancy-based baseline. All IRL-based methods resulted in a similar number of interventions, but exhibited some qualitatively different navigation behaviors. Figure \ref{fig:qualitative_nav} highlights several key differences in observed behaviors.

\begin{table}[]
    \centering
    \begin{tabular}{c||c|c|c}
         Method & Auto. Dist. & Auto. Speed & Interventions \\
         \hline
         Baseline & $1465m$ & $3.21m/s$ & 7 \\
         IRL (CVaR -0.9) & $1482m$ & $3.25m/s$ & 4 \\
         IRL (CVaR 0.0) & $1498m$ & $3.29m/s$ & \textbf{3} \\
         IRL (CVaR 0.9) & $1592m$ & $3.14m/s$ & 4 \\
    \end{tabular}
    \caption{Navigation metrics for each method. Distance and speed columns are not bolded as higher/lower values don't necessarily mean better performance.}
    \vspace{-0.5cm}
    \label{tab:intervention_results}
\end{table}


Overall, we are again able to observe significant improvement over the baseline when using IRL-based costmaps. All IRL-based methods had roughly the same number of interventions, but exhibited different navigation behaviors, especially with respect to tall grass. In general, more conservative settings of CVaR ($>0.8$), caused the ATV to avoid patches of moderate-height grass when possible, while more aggressive settings of CVaR ($<-0.8$), caused the ATV to drive through taller patches of grass. As one would expect, average distance traveled increased with $\text{CVaR}_\nu$, as there were monotonically more high-cost regions to avoid. The occupancy-based method traveled the shortest amount of autonomous distance as a result of having more human interventions. Interestingly, we observed a significant difference between the speeds and distance traveled for $\text{CVaR}_{-0.9}$ and $\text{CVaR}_0$ were relatively small, while $\text{CVaR}_{0.9}$ was considerably slower and traveled farther. 


\section{Conclusion and Future Work}

We present an efficient method of training risk-aware costmaps with MEDIRL and demonstrate improved performance over occupancy-based baselines in challenging navigation scenarios. We also show that changing the risk-tolerance of the vehicle via CVaR results in different behaviors with respect to uncertain terrains such as tall grass.

We believe that there are three important extensions of this work. First, improving the set of features that MEDIRL can use should in theory lead to improvement. To that end, we are interested in adding visual features in addition to our lidar features. Methods such as that presented by Harley et al. \cite{harley2022simple} are immediately applicable to our current feature representation, and would allow us to learn off of features such as RGB, semantic segmentation or even deep features. Second, we believe that significant navigation improvements can come from considering the velocity dimension. It follows intuition that different terrains can incur different costs depending on speed. As such, we believe that improvement in navigation can be achieved by predicting costmaps with an additional velocity dimension. Finally, while we demonstrate that different CVaR values result in different navigation behaviors, deciding the correct CVaR value remains an open problem. It seems that there are many factors that influence the level of risk an ATV should take, including speed, surroundings and operator preference. As such, we are also interested in designing controllers that are able to adapt their risk-tolerance online.

\newpage

\appendix

\subsection{Detailed Algorithms}

We present in detail the algorithms described in the main paper. Algorithm \ref{algo:mppi_state_visitations} describes the algorithm for extracting state visitations from an MPPI solution. Algorithm \ref{algo:medirl_train} describes the training step for our algorithm. Algorithm \ref{algo:medirl_test} describes the risk-aware costmap generation process.

\begin{algorithm}
\DontPrintSemicolon
\caption{Computation of State Visitation Frequencies (SVF) from Weighted Trajectories}
\label{algo:mppi_state_visitations}
\KwIn{Initial State $x_0$, Weights $\boldsymbol{\eta}$, Trajectories $\boldsymbol{\tau}$, Map resolution $M_{res}$}
\KwOut{State Visitation Frequencies $D$}

\For{$\tau_n, \eta_n \in \boldsymbol{\tau}, \boldsymbol{\eta}$}{
    \For{$x_t \in \tau_n$} {
        $i, j \leftarrow \left \lfloor {p(x_t) / M_{res}} \right \rfloor$ \;
        $D[i, j] \leftarrow D[i, j] + \eta_n$ \;
    }
}

$D  \leftarrow \frac{D}{\sum_{i, j} D[i, j]}$ \;

return $D$ \;

\end{algorithm}

\begin{algorithm}
\DontPrintSemicolon
\caption{Training Step for Fast MEDIRL}
\label{algo:medirl_train}
\KwIn{\\
        \quad Dataset $\mathcal{D}$ of expert trajectories $\boldsymbol{\tau_E}$, \\
        \quad Map features $\mathbf{M}$ , \\
        \quad Goal weight $\kappa$, \\
        \quad FCN Ensemble $F_\theta(M) : \mathcal{R}^{W \times H \times D} \rightarrow \mathcal{R}^{B \times W \times H}$, \\
        \quad MPPI $MPPI(x_s, x_g, C, \lambda): \mathcal{X} \rightarrow (\eta_n, \tau_n) \times N$}

\While{not converged}
{
    $\tau^E, M \sim \mathcal{D}$ \hfill $\triangleleft$ Sample from dataset \;
    $x_0 = \tau^E_0, x_g = \tau^E_{t-1}$ \hfill $\triangleleft$ set start/goal \;
    $f_\theta \sim \boldsymbol{F_\theta}$ \hfill $\triangleleft$ sample FCN from ensemble \;
    $C = f_\theta(M_0)$ \hfill $\triangleleft$ Compute costmap from FCN \;
    $\boldsymbol{\tau^L}, \boldsymbol{\eta^L} = MPPI(x_0, x_g, C, \kappa)$ \;
    $D^E = SVF(\tau_E)$ \hfill $\triangleleft$ Compute state visitations \;
    $D^L = SVF(\tau^L, \eta^L)$ \quad via Algorithm \ref{algo:mppi_state_visitations}\;
    $\nabla_z J = D^E - D^L$ \hfill $\triangleleft$ Gradient via \cite{wulfmeier2015deep} \;
    backprop($\nabla_z J, f_\theta$) \hfill $\triangleleft$ Update FCN grads\;
    $\theta \leftarrow \text{Adam}(\theta - \nabla_\theta J)$ \hfill $\triangleleft$ Update via \cite{kingma2014adam}\;
}
\end{algorithm}

\begin{algorithm}
\DontPrintSemicolon
\caption{Inference Step for Fast MEDIRL}
\label{algo:medirl_test}
\KwIn{\\
        \quad Map features $M$, \\
        \quad FCN Ensemble $F_\theta(M) : \mathcal{R}^{W \times H \times D} \rightarrow \mathcal{R}^{B \times W \times H}$, \\
        \quad Risk level $\nu$}

$C_\nu = \boldsymbol{0}^{m \times n}$ \hfill $\triangleleft$ Initialize empty costmap\;
$\boldsymbol{C} = \boldsymbol{F_\theta}(M)$ \hfill $\triangleleft$ \hfill $\triangleleft$ Create costmap for each FCN\;
\For{$i,j \in ([0 \hdots m] \times [0 \hdots n])$}
{
    $\boldsymbol{c} = \boldsymbol{C}[:, i, j]$ \hfill $\triangleleft$ Get each FCN's cell output \;
    $C_\nu[i, j] = CVaR_\nu(\boldsymbol{c})$ \hfill $\triangleleft$ Compute CVaR \;
}

return $C_\nu$

\end{algorithm}

\subsection{Lidar Mapping Algorithm}

Our lidar mapping algorithm is presented in detail in Algorithm \ref{algo:lidar_mapping}. 

\begin{algorithm*}
\DontPrintSemicolon
\caption{Lidar Mapping Algorithm}
\label{algo:lidar_mapping}
\KwIn{Buffer of registered pointclouds $P_b$, pointcloud skip $k_p$, Map origin $(o_x, o_y)$, Map size $(l_x, l_y)$, Map resolution $r$, overhang limit $k_{overhang}$}
\KwOut{Terrain feature tensor $X$}

$n_x = \lfloor \frac{l_x}{r} \rfloor$ \;
$n_y = \lfloor \frac{l_y}{r} \rfloor$ \;
$X = \mathbf{0} ^ {n_x \times n_y \times 12}$ \hfill $\triangleleft$ Initialize map tensor \;
$P = \sum_{i=0}^{|P_b| / k_p} P_{i * k_p}$  \hfill $\triangleleft$ Aggregate pointclouds from buffer\;

\For{$i=0 \hdots n_x$}{
    \For{$j=0 \hdots n_y$}{
        $P_m = \{p, \forall p \in P | (\frac{P_x - o_x}{r} = i) \land (\frac{P_y - o_y}{r} = j)\}$  \hfill $\triangleleft$ Get all points in a given column \;
        
        $X[i, j, 0] = min_{p \in P_m} [p_z]$ \hfill $\triangleleft$ Get min height \;
        $X[i, j, 1] = max_{p \in P_m} [p_z]$  \hfill $\triangleleft$ Get max height \;
    }
}

$X[:, :, 4] = G * inflate(X[:, :, 0])$ \hfill $\triangleleft$ Generate terrain estimate by inflating and low-pass filtering min height \;
$X[:, :, 5] = S_x * |X[:, :, 2]| + S_y * |X[:, :, 2]|$ \hfill $\triangleleft$ Get terrain slope via derivative filter\;

\For{$i=0 \hdots n_x$}{
    \For{$j=0 \hdots n_y$}{
        $P_m = \{p, \forall p \in P | (\frac{p_x - o_x}{r} = i) \land (\frac{p_y - o_y}{r} = j) \land (p_z < X[i, j, 2] + k_{overhang})\}$ \hfill $\triangleleft$ Filter overhanging points \;
        
        $X[i, j, 2] = \max_z [p_z \forall p \in P_m]$ \hfill $\triangleleft$ Get the max height of the cell, saturating at the overhang limit\;

        $X[i, j, 3] = \frac{1}{|P_m|} \sum_{p \in P_m} [p_z]$ \hfill $\triangleleft$ Get the mean height of the cell\;

        $X[i, j, 6] = X[i, j, 2] - X[i, j, 4]$ \hfill $\triangleleft$ Get the height of the cell relative to terrain \$

        $\lambda_1, \lambda_2, \lambda_3 = SVD(P_m)$ \hfill $\triangleleft$ Get the SVD decomposition of the cell points \;
        $X[i, j, 7] = \frac{\lambda_1 - \lambda_2}{\lambda_1}$ \hfill $\triangleleft$ Get SVD1 \;
        $X[i, j, 8] = \frac{\lambda_2 - \lambda_3}{\lambda_1}$ \hfill $\triangleleft$ Get SVD2 \;
        $X[i, j, 9] = \frac{\lambda_3}{\lambda_1}$ \hfill $\triangleleft$ Get SVD3 \;
        $X[i, j, 10] = \frac{\lambda_3}{\lambda_1 + \lambda_2 + \lambda_3}$ \hfill $\triangleleft$ Get roughness \;
        $X[i, j, 11] = \mathbf{1}[|P_m| = 0]$ \hfill $\triangleleft$ Get unknown \;
    }
}

return $X$ \;
\end{algorithm*}

\subsection{Navigation Course Overview}

\begin{figure}
    \centering
    \includegraphics[width=\linewidth]{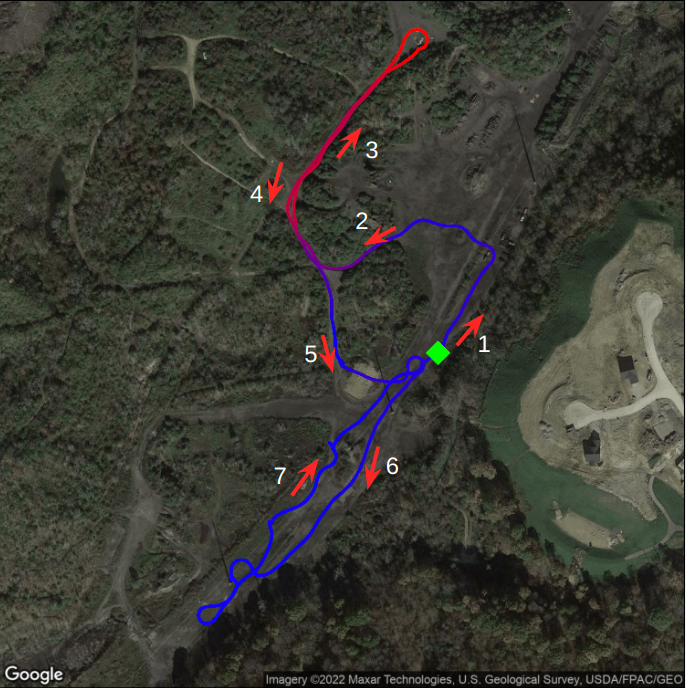}
    \caption{An overview of our navigation course. In total, the course was roughly 1.6km, and included several challenging scenarios such as going over slopes and through tall grass. The course began and ended at the green diamond, and followed the red arrows. The path is colored according to elevation change.}
    \label{fig:nav_course}
\end{figure}

A more detailed description of the course used for the navigation experiments and representative terrains is presented in Figure \ref{fig:nav_course}.

\subsection{Comparison of CVaR Plots}

\begin{figure*}
     \centering
     \begin{subfigure}[t]{\linewidth}
         \centering
         \includegraphics[scale=0.4]{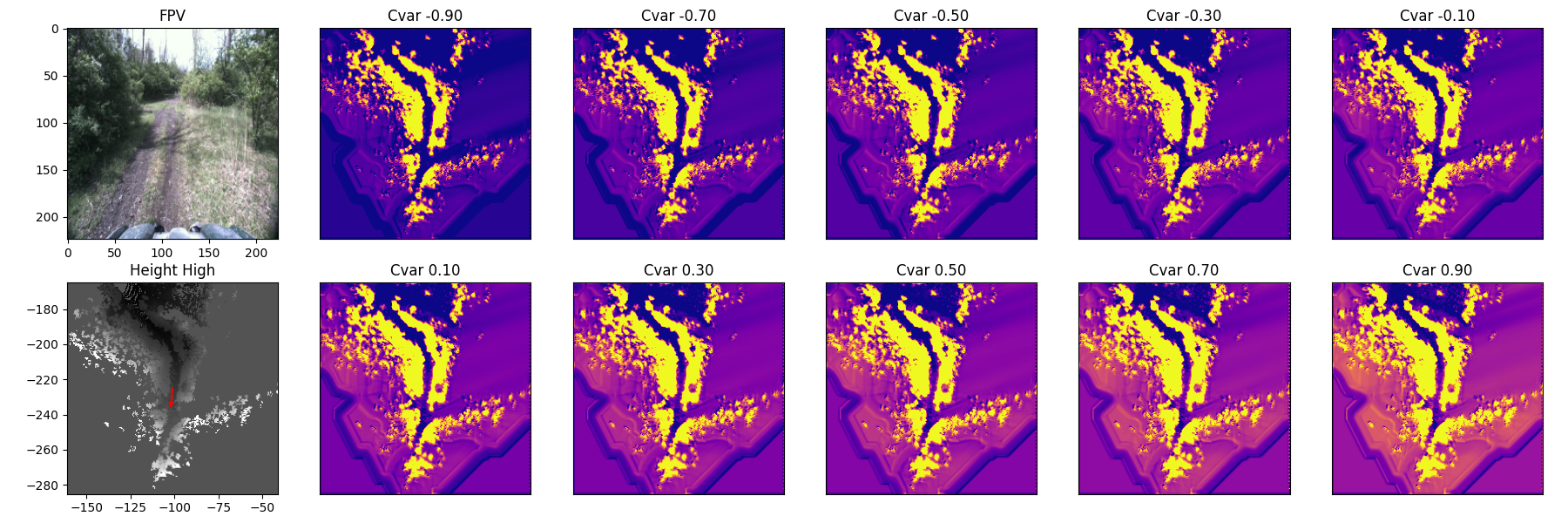}
         \caption{Corridor Scenario. The relative cost of obstacles and trail are consistent for all $\text{CVaR}_\nu$.}
     \end{subfigure}
     \hfill
     \begin{subfigure}[t]{\linewidth}
         \centering
         \includegraphics[scale=0.4]{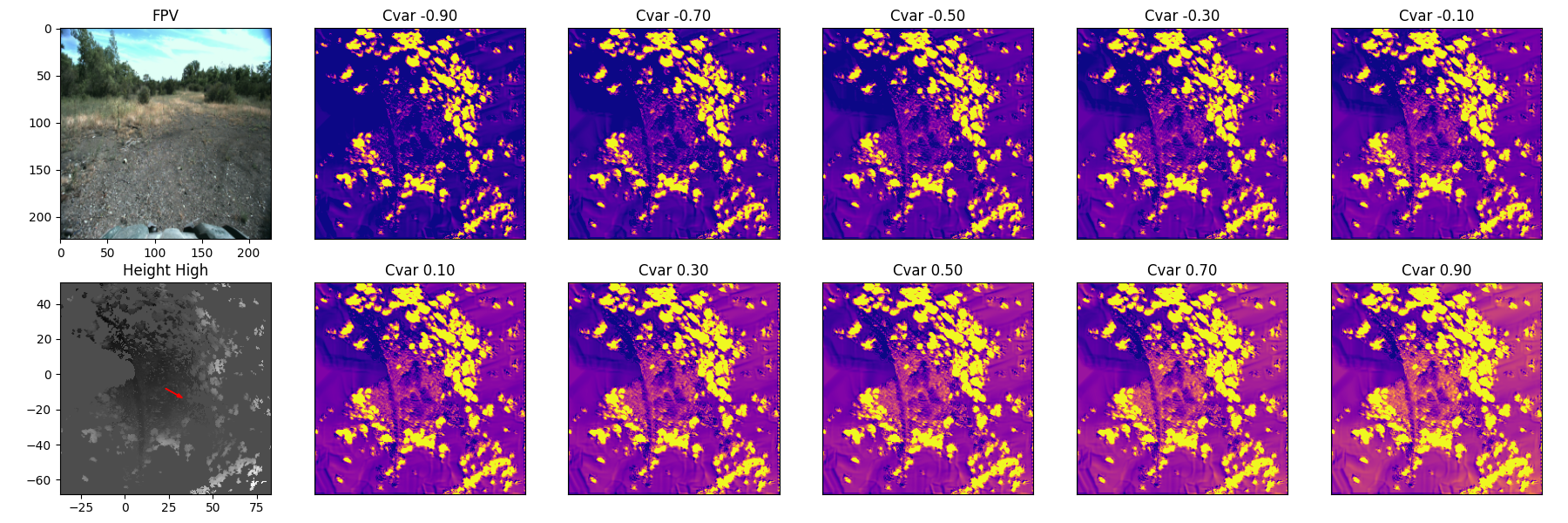}
         \caption{Tall Grass Scenario. The relative cost of grass increases with $\text{CVaR}_\nu$.}
     \end{subfigure}
     \hfill
     \begin{subfigure}[t]{\linewidth}
         \centering
         \includegraphics[scale=0.4]{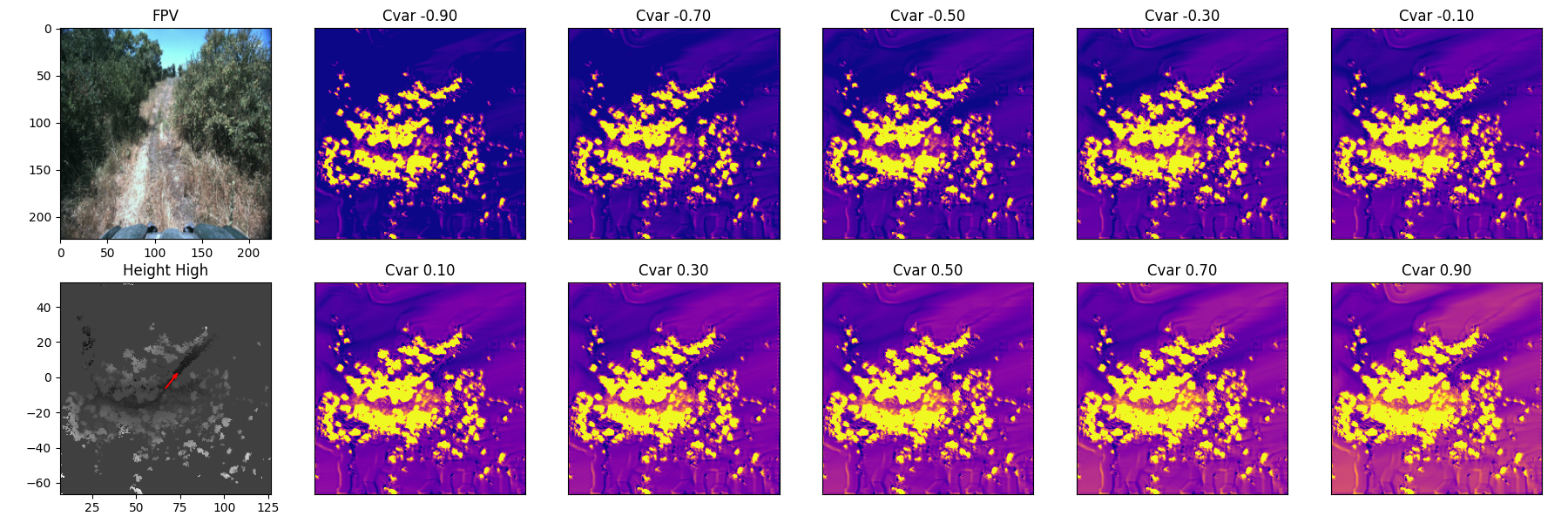}
         \caption{Slope Scenario. The relative cost of the slope in the FPV increases with $\text{CVaR}_\nu$.}
     \end{subfigure}
     \hfill
     \caption{Resnet Results on representative terrains in the test set. Note that in uncertain regions such as grass and slopes, the resnet-based costmap adjusts costs based on CVaR.}
     \label{fig:resnet_cvar_qual}
\end{figure*}

\begin{figure*}
     \centering
     \begin{subfigure}[t]{\linewidth}
         \centering
         \includegraphics[scale=0.4]{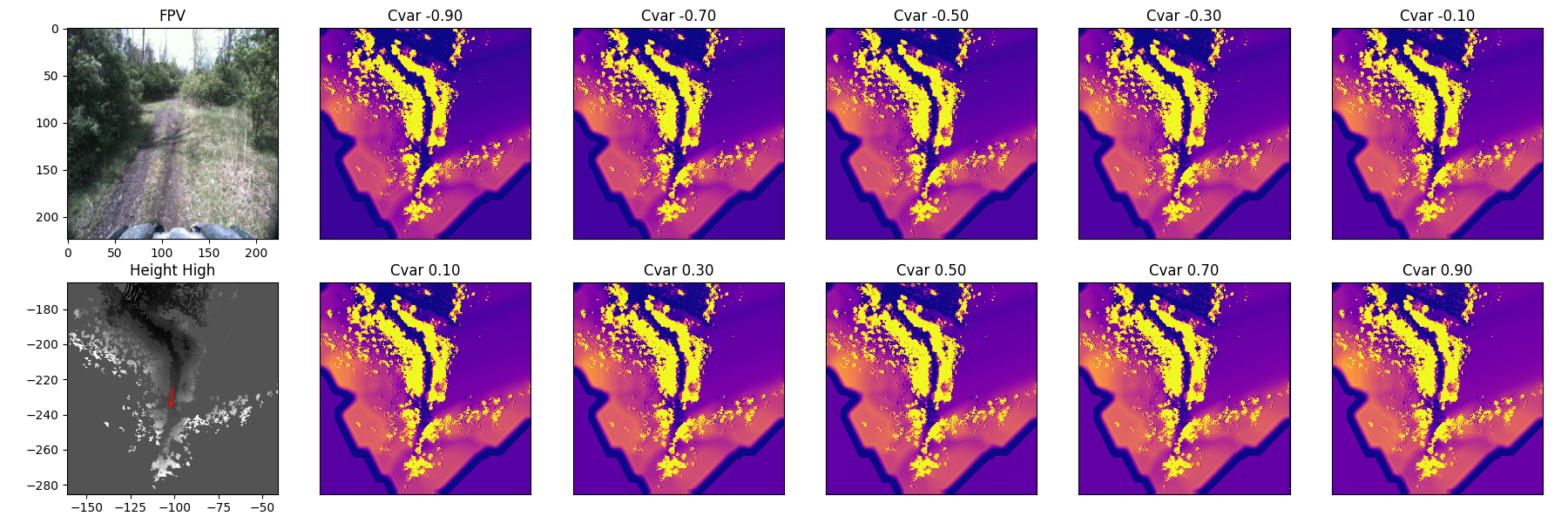}
         \caption{Corridor Scenario}
     \end{subfigure}
     \hfill
     \begin{subfigure}[t]{\linewidth}
         \centering
         \includegraphics[scale=0.4]{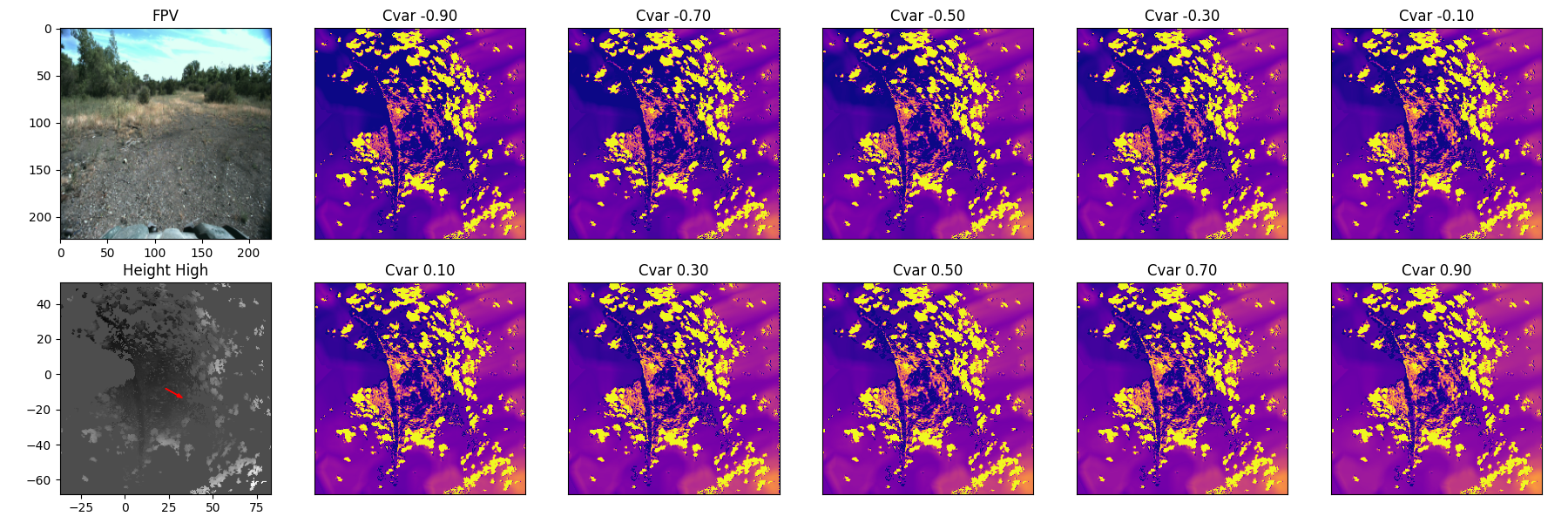}
         \caption{Tall Grass}
     \end{subfigure}
     \hfill
     \begin{subfigure}[t]{\linewidth}
         \centering
         \includegraphics[scale=0.4]{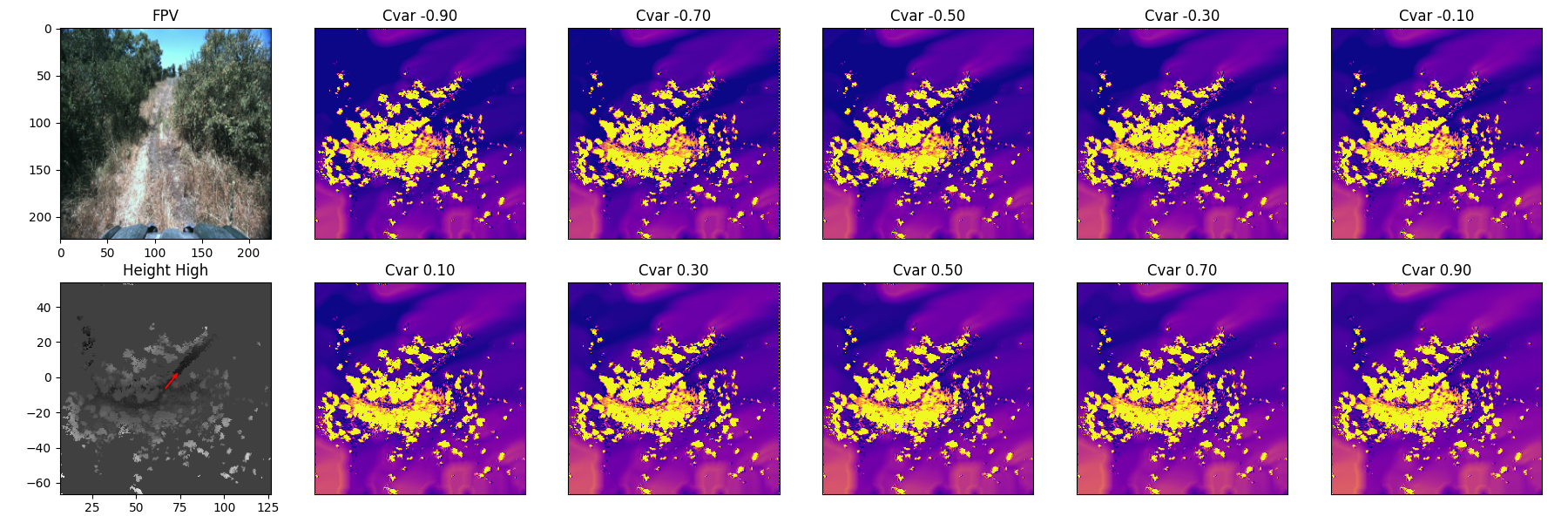}
         \caption{Slope Scenario}
     \end{subfigure}
     \hfill
     \caption{Linear Results on representative terrains in the test set. Note that the costmaps are more or less unchanged with respect to CVaR.}
     \label{fig:linear_cvar_qual}
\end{figure*}

Presented below are additional qualitative visualizations of the costmaps learned from linear features and resnets. Visualizations are produced from representative examples of terrain from the test set in Figures \ref{fig:resnet_cvar_qual} and \ref{fig:linear_cvar_qual}. Note that the colors in these figures  are normalized for \textit{each} setting of CVaR. That is, changes in color denote changes in cost \textit{relative} to other costs in that particular subplot. Also note that the vehicle is located in the center of each BEV plot, and the orientation of the vehicle in the map is notated via the red arrow.

\subsection{KBM Dynamics}
The full form of our KBM dynamics are presented in Equation \ref{eq:kbm}, with hyperparameters in Table \ref{tab:model_params}.

\begin{equation}
    \label{eq:kbm}
    \mathbf{X} = 
    \begin{bmatrix}
        x \\
        y \\
        \theta \\
        v \\
        \delta \\
    \end{bmatrix},
    \mathbf{U} =
    \begin{bmatrix}
        v_{target} \\
        \delta_{target}
    \end{bmatrix},
    \mathbf{\dot{X}} = 
    \begin{bmatrix}
        v cos(\theta) \\
        v sin(\theta) \\
        v \frac{tan(\delta)}{L} \\
        K_v (v_{target} - v) \\
        K_\delta (\delta_{target} - \delta) \\
    \end{bmatrix}
\end{equation}

\subsection{Hyperparameters}
The hyperparameters used in our experiment are provided in Tables \ref{tab:model_params} and \ref{tab:mppi_params}.

\begin{table}[]
    \centering
    \begin{tabular}{c|c}
        Parameter & Value \\
        \hline
         $L$ & $3.0m$ \\
         $K_v$ & $1.0$ \\
         $K_\delta$ & $10.0$ \\
         $v$ limits (IRL) & $[2.0, 15.0] m/s$ \\
         $v$ limits (MPC)& $[1.5, 3.5] m/s$ \\
         $\delta$ limits& $[-0.52, 0.52] rad$ \\
         $\omega$ limits& $[-.0.2, 0.2] rad/s$ \\
         $dt$(IRL) & $0.1s$ \\
         $dt$(MPC) & $0.15s$ \\
    \end{tabular}
    \caption{Table of Parameter Values for ATV}
    \label{tab:model_params}
\end{table}

\begin{table}[]
    \centering
    \begin{tabular}{c|c}
        Parameter & Value \\
        \hline
        Iterations / step (IRL) & 10 \\
        Iterations / step (MPC) & 1 \\
        $H$(IRL) & 75 \\
        $H$(MPC) & 60 \\
        $N$(IRL) & 2048 \\
        $N$(MPC) & 512 \\
        $\kappa$(IRL) & 20 \\
        $\kappa$(MPC) & 10 \\
        $\Sigma$ & $diag([1.0, 0.1])$ \\
        $\lambda$ & 20 \\
        $\alpha$ & 0.9 \\
        $K$ & 10.0 \\
        $dK$ & 5.0 \\
    \end{tabular}
    \caption{MPPI parameters}
    \label{tab:mppi_params}
\end{table}

\subsection{Comparison of Dataset Difficulty}

\begin{figure*}
     \centering
     \begin{subfigure}[t]{\linewidth}
         \centering
         \includegraphics[scale=0.4]{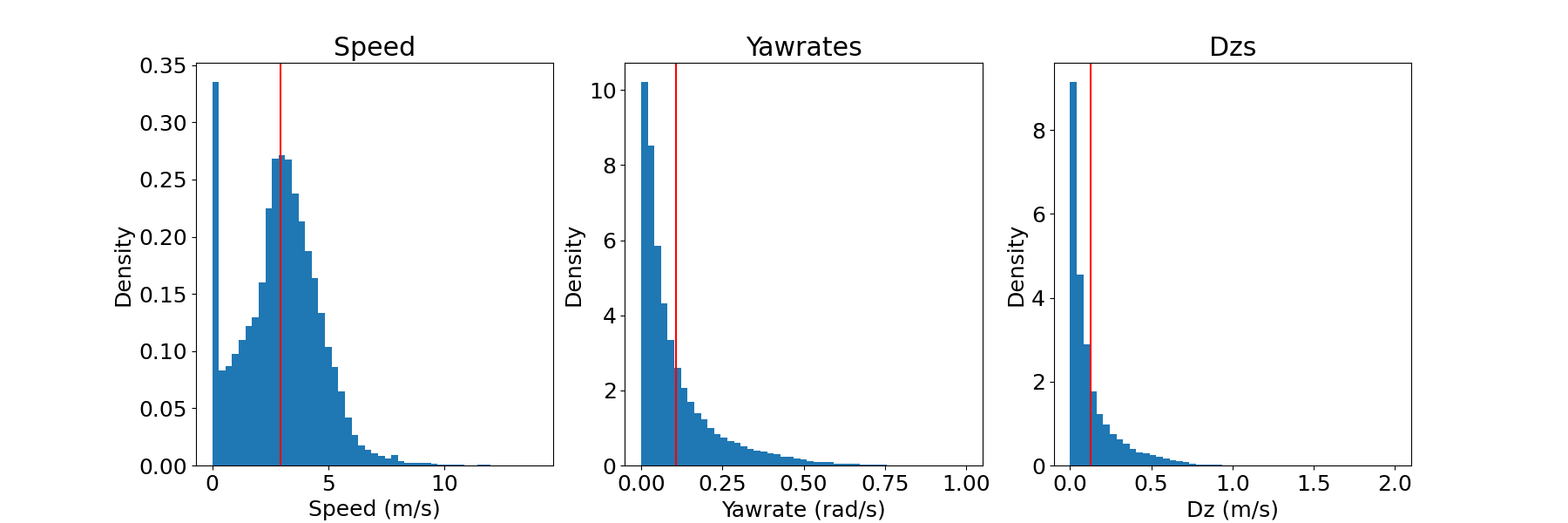}
         \caption{Tartandrive difficulty}
     \end{subfigure}
     \hfill
     \begin{subfigure}[t]{\linewidth}
         \centering
         \includegraphics[scale=0.4]{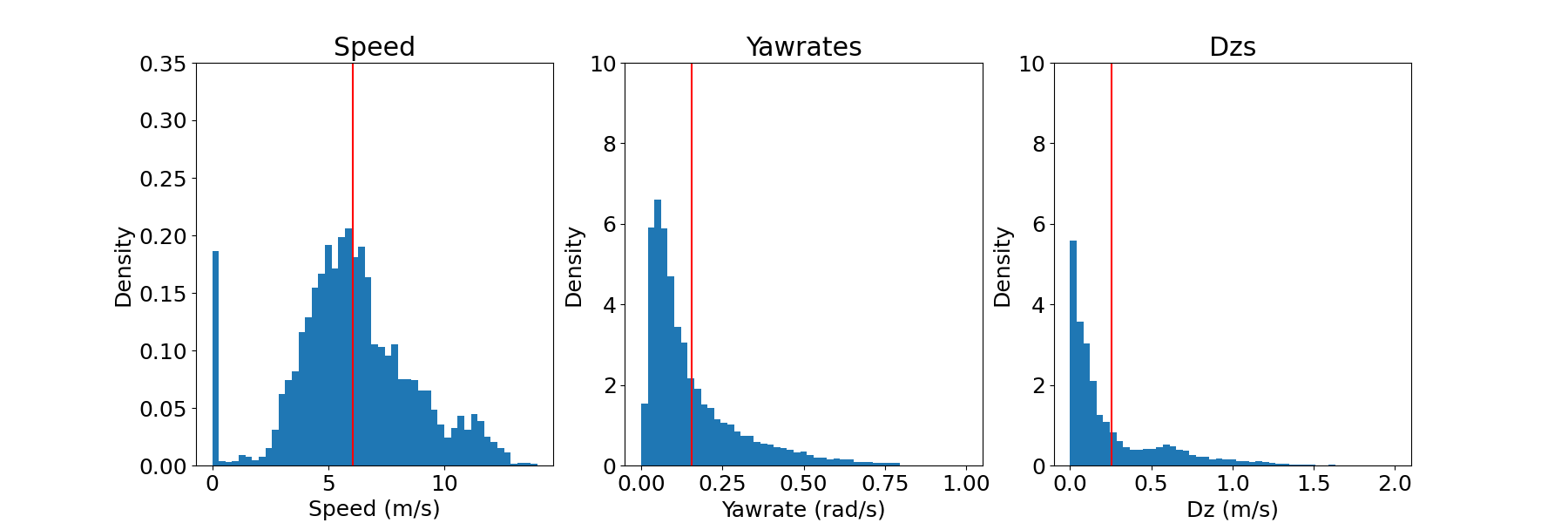}
         \caption{IRL dataset difficulty}
     \end{subfigure}
     \hfill
     \caption{Comparison of dataset difficulties between our dataset and TartanDrive. Our IRL dataset has higher mean difficulty and a wider distribution of difficulty, as well.}
     \label{fig:dataset_compare}
\end{figure*}

We present a comparison of our dataset to TartanDrive \cite{triest2022tartandrive}. As mentioned in the main paper, the biggest difference in the datasets is the availability of lidar in our IRL dataset. This allows us to generate maps that are $40m$ in the vehicle's forward direction, as opposed to Tartandrive's $10m$. As a result, we are able to collect a dataset that contains much more aggressive driving. We quantify this in Figure \ref{fig:dataset_compare}, where we report the distribution of three quantities associated with driving aggression; speed, yaw rate and integrated change in height. We can observe that on our IRL dataset exceeds TartanDrive on all quantities, especially speed. Furthermore, the distributions in the IRL dataset appear to be more long-tailed.

{
\bibliographystyle{IEEEtran}
\bibliography{refs}
}

\end{document}